%% file: main_arxiv.tex
\documentclass[runningheads]{llncs}
\usepackage[T1]{fontenc}
\usepackage{amsmath,amssymb,amsfonts}
\usepackage{algorithmic}
\usepackage{graphicx}
\usepackage{textcomp}
\usepackage{xcolor}
\usepackage{algorithm}
\usepackage{booktabs}
\usepackage{microtype}
\begin{document}
\title{Weakly Supervised Pixel-Level Annotation with Visual Interpretability}
%
%
\author{Basma Nasir\inst{1}\orcidID{0009-0003-1589-8722} \and
Tehseen Zia\inst{1}\orcidID{0000-0001-8176-3373} \and
Muhammad Nawaz\inst{2}\orcidID{0009-0001-8404-7695} \and 
Catarina Moreira\inst{2}\orcidID{0000-0002-8826-5163}}
\authorrunning{Basma et al.}
%
\institute{Department of Computer Science, COMSATS University Islamabad, Pakistan
\email{sp22-rcs-004@isbstudent.comsats.edu.pk}, \email{tehseen zia@comsats.edu.pk}\\
 \and
Data Science Institute, University of Technology Sydney, Australia\\
\email{muhammad.nawaz@student.uts.edu.au},
\email{catarina.pintomoreira@uts.edu.au}}
\maketitle              
\begin{abstract}
Medical image annotation is essential for diagnosing diseases, yet manual annotation is time-consuming, costly, and prone to variability among experts. To address these challenges, we propose an automated explainable annotation system that integrates ensemble learning, visual explainability, and uncertainty quantification. Our approach combines three pre-trained deep learning models—ResNet50, EfficientNet, and DenseNet—enhanced with XGrad-CAM for visual explanations and Monte Carlo Dropout for uncertainty quantification. This ensemble mimics the consensus of multiple radiologists by intersecting saliency maps from models that agree on the diagnosis while uncertain predictions are flagged for human review. We evaluated our system using the TBX11K medical imaging dataset and a Fire segmentation dataset, demonstrating its robustness across different domains. Experimental results show that our method outperforms baseline models, achieving 93.04\% accuracy on TBX11K and 96.4\% accuracy on the Fire dataset. Moreover, our model produces precise pixel-level annotations despite being trained with only image-level labels, achieving Intersection over Union IoU scores of 36.07\% and 64.7\%, respectively. By enhancing the accuracy and interpretability of image annotations, our approach offers a reliable and transparent solution for medical diagnostics and other image analysis tasks.

\end{abstract}

\section{Introduction}

Medical image annotation, also referred to as data tagging or labeling, is essential for identifying pathological conditions and guiding clinical decision-making. This process involves adding descriptive labels to medical images from modalities such as Magnetic Resonance Imaging (MRI) and Computed Tomography (CT) scans, enabling accurate detection of anatomical structures, tumors, and other abnormalities. Traditionally, this task is performed manually by radiologists, whose expertise ensures diagnostic precision. However, manual annotation is both time-consuming and resource-intensive, especially for large datasets, and is susceptible to inter-observer variability due to differences in expert judgment \cite{Rahimi21}. These limitations can delay diagnosis and compromise the consistency of medical image analysis, particularly in regions with a shortage of skilled professionals, such as Pakistan \cite{ref3}.

Recent advancements in Artificial Intelligence (AI), particularly deep learning, have transformed medical image analysis by enabling automated annotation systems \cite{galbusera2024image,hsieh2023mdf,wang2024comprehensive,balasamy2024medical}. These systems can significantly reduce the time and cost associated with manual labeling while improving the accuracy and consistency of annotations \cite{ref5}. Despite their promise, AI models often function as "black boxes," offering limited insight into their decision-making processes \cite{alzubaidi2023towards}. This lack of transparency limits their acceptance in clinical practice, where understanding the reasoning behind predictions is essential for building trust among healthcare professionals \cite{ref6,velmurugan2021developing}. To address this, the field of Explainable Artificial Intelligence (XAI) has emerged, focusing on developing interpretable models that provide visual and conceptual explanations for their outputs \cite{ref7,neves2024shedding,moreira2025benchmarking}.

However, existing explainable models primarily rely on datasets annotated with pixel-level masks, which are costly and labor-intensive to produce \cite{ref10}. Furthermore, most AI systems struggle with the open-set problem, where they encounter previously unseen data that falls outside the scope of their training set \cite{ref33}. The ability to detect and flag such cases is critical for ensuring diagnostic reliability. Therefore, there is a need for an automated annotation system that can generate accurate pixel-level masks using only image-level labels while also estimating prediction uncertainty to identify novel data.

To bridge this gap, we propose an automated explainable annotation system that integrates ensemble learning, visual explainability, and uncertainty quantification. Our method combines three pre-trained deep learning models, ResNet50, EfficientNet, and DenseNet, augmented with XGrad-CAM to generate saliency maps that highlight class-specific image regions. Monte Carlo Dropout is incorporated into each model to quantify prediction uncertainty, allowing the system to flag ambiguous cases for human review. By intersecting the saliency maps of models that agree on the classification, the system produces pixel-level annotations that closely align with expert-labeled ground truth masks. This approach eliminates the need for pixel-level annotations during training, significantly reducing the time and resources required for model development.

The main contributions of this paper are as follows:

\begin{itemize}
    \item \textbf{Ensemble Learning for Robust Predictions:} We introduce an ensemble framework combining ResNet50, EfficientNet, and DenseNet to enhance the accuracy, reliability, and consistency of medical image annotations.

    \item \textbf{Weakly Supervised Pixel-Level Annotations:} We extend XGrad-CAM beyond standard visual explanations, enabling pixel-level mask generation using only image-level labels, significantly reducing the need for manual annotations.  

    \item \textbf{Uncertainty Quantification for Open-Set Detection:} By integrating Monte Carlo Dropout, our system estimates prediction uncertainty, allowing it to flag ambiguous and novel data, improving reliability in real-world scenarios.  
\end{itemize}

By addressing the limitations of traditional AI models and introducing an interpretable, reliable, and resource-efficient annotation system, our work advances medical image analysis and the broader field of Explainable Artificial Intelligence.

\section{Literature Review}

Recent advancements in computer vision and explainable AI have significantly improved object segmentation and interpretation in medical and general imaging. However, many state-of-the-art methods still rely on pixel-level annotations, which are costly and time-consuming \cite{Rahimi21}. Additionally, neural networks often lack transparency, limiting their applicability in high-stakes domains like healthcare \cite{alzubaidi2023towards}. This section reviews key developments in semantic segmentation, weakly supervised learning, generative adversarial networks, and uncertainty quantification, highlighting their strengths and limitations. The identified research gaps provide the foundation for our proposed approach, which integrates ensemble learning, XGrad-CAM explainability, and Monte Carlo Dropout-based uncertainty estimation to generate reliable pixel-level annotations using only image-level labels.

\subsection{Semantic Segmentation}

Recent advances in semantic segmentation have improved the accuracy of object localization in medical imaging. For example, GroupViT \cite{ref12} employs a Hierarchical Grouping Vision Transformer that segments image regions using text supervision. While effective, its reliance on pre-trained object detectors with bounding box annotations limits its applicability in scenarios where pixel-level annotations are unavailable. In medical imaging, SMR-UNet \cite{ref13} uses self-attention mechanisms, multi-scale feature integration, and residual structures to improve the segmentation of lung nodules. Although it captures both local and global contextual information, its performance depends on detailed pixel-level annotations, which are costly to obtain.

USegTransformer-P and USegTransformer-S \cite{ref14} combine transformers with CNNs to improve segmentation accuracy. However, both models require large, annotated datasets and high computational resources, making them less practical in data-limited or resource-constrained environments. Similarly, the hybrid attention-based residual UNet \cite{ref15} enhances brain tumor segmentation but relies on resized inputs, which may omit diagnostic information critical for clinical applications.

While these methods have advanced pixel-level segmentation, they rely heavily on pixel-level annotations and are sensitive to data quality and computational constraints. In contrast, our approach addresses these limitations by using a weakly supervised learning framework that requires only image-level labels. By integrating XGrad-CAM within an ensemble of ResNet50, EfficientNet, and DenseNet, our method produces pixel-level masks without manual annotations, reducing data preparation costs while maintaining interpretability and reliability.

\subsection{Weakly Supervised Semantic Segmentation}

Weakly supervised semantic segmentation (WSSS) aims to segment objects or regions of interest without relying on detailed pixel-level annotations. This is particularly important in medical imaging, where obtaining pixel-level labels is time-consuming and resource-intensive.

The Multi-class Token Transformer (MCTformer) \cite{ref16} enhances object localization using class-specific attention mechanisms, improving segmentation precision. However, it still requires class labels and struggles with complex boundaries, limiting its applicability in medical contexts. Causal Class Activation Maps (C-CAM) \cite{ref17} address WSSS challenges in medical imaging by leveraging anatomy and co-occurrence causalities, generating pseudo-segmentation masks with clearer boundaries. Yet, C-CAM’s reliance on heatmap thresholding can reduce accuracy for small or overlapping regions.

ReFit \cite{ref18} integrates unsupervised segmentation and saliency methods to create edge maps that refine object boundaries using Grad-CAM. Despite improved boundary delineation, its effectiveness depends on the quality of edge maps, which can be inconsistent in noisy medical images. For brain tumour segmentation, \cite{ref19} proposes classifiers trained with only image-level labels, producing heatmaps that guide ROI segmentation. However, its performance is limited by the thresholding process, which may exclude subtle features. A 3D segmentation technique \cite{ref20} combines semi-supervised and self-supervised learning to generate pseudo-labels, but its reliance on ground truth labels for central slices restricts its use in datasets with specific annotations.

Although these methods reduce annotation requirements, they still depend on partial pixel-level labels or struggle with accurate segmentation from image-level annotations. Our approach overcomes these limitations by generating pixel-level masks using only image-level labels. By integrating XGrad-CAM within an ensemble of ResNet50, EfficientNet, and DenseNet, we produce accurate annotations without manual pixel-level labeling. Additionally, using Monte Carlo Dropout quantifies uncertainty, ensuring reliable segmentation even in open-set scenarios where unseen data may occur.

\subsection{Generative Adversarial Networks and Interpretability}

Generative Adversarial Networks (GANs) \cite{ref21} consists of two components: a generator that creates synthetic data and a discriminator that distinguishes between real and generated data. GANs have been applied across various domains, including realistic image synthesis \cite{ref22}, domain translation, and medical image generation \cite{ref23}. However, their limited explainability hinders their use in critical applications like medical imaging. Existing explainability methods for GANs, such as image-to-image translation \cite{ref24,ref25}, struggle to generate accurate pixel-level masks, reducing their effectiveness in visualizing decision-making processes.

CycleGAN \cite{ref26}, designed for unpaired image-to-image translation, has improved tissue segmentation and disease detection in cardiac, liver, and retinal imaging \cite{ref27}. Although it enhances visual interpretability by revealing disease impacts, it does not produce binary masks, limiting its use for precise annotations. Class Activation Maps (CAM) \cite{ref28} provide visual explanations by highlighting class-specific image regions using the final convolutional layer. Grad-CAM \cite{ref29} extends this approach by leveraging gradient information to generate coarse localization maps. However, CAM and Grad-CAM only offer heatmaps and cannot produce binary masks, limiting their utility for pixel-level annotation.

MDVA-GAN \cite{ref30} addresses this limitation by integrating CycleGAN with Grad-CAM to visualize multi-class features and generate binary masks. Despite this improvement, MDVA-GAN often misclassifies images from unseen classes, reducing its reliability in open-set scenarios. Additionally, its reliance on adversarial training can lead to unstable results, especially when training data is limited.

These limitations highlight the need for an approach to generate pixel-level annotations from image-level labels while maintaining reliability in open-set scenarios. Our method addresses these challenges by integrating XGrad-CAM within an ensemble of ResNet50, EfficientNet, and DenseNet to generate pixel-level masks without adversarial training. Unlike MDVA-GAN, our approach quantifies prediction uncertainty using Monte Carlo Dropout, enabling it to flag ambiguous or novel data, ensuring more reliable annotations even when encountering unseen classes.

\subsection{Uncertainty Quantification}

Deep learning has advanced diagnostic evaluations in medical imaging, including CT, MRI, ultrasound, and histopathology \cite{ref31}. Despite these advancements, neural networks often function as "black boxes," offering limited insight into their decision-making processes \cite{ref32}. This opacity raises safety and reliability concerns, as models can overestimate their confidence when processing anomalous data \cite{ref33} and are vulnerable to adversarial attacks \cite{ref34}. Identifying these limitations is critical for ensuring the reliable integration of DL models into clinical workflows.

Uncertainty Quantification (UQ) techniques address these challenges by estimating the confidence of model predictions, enabling the identification of ambiguous cases that require human review \cite{ref36}. This is especially important in healthcare, where undetected errors can lead to misdiagnoses and inappropriate treatments \cite{ref37}. Traditional neural networks use the Softmax function to output probability distributions across classes, but these probabilities often do not reflect true uncertainty, particularly in open-set scenarios where the input data differs from the training distribution.

The Uncertainty-Inspired Open Set (UIOS) model \cite{ref38} improves the detection and classification of retinal anomalies by using evidential uncertainty estimation. While UIOS is trainable and computationally efficient, its reliance on specific annotations, such as central slice labels, and the need for manual threshold tuning limit its applicability to datasets with predefined reference points. This dependency restricts its generalizability to broader medical imaging tasks.

Our approach overcomes these limitations by integrating Monte Carlo Dropout into an ensemble of ResNet50, EfficientNet, and DenseNet models. Unlike UIOS, our method does not require specific annotations or manual thresholding. By performing multiple stochastic forward passes during inference, Monte Carlo Dropout estimates the variance of predictions, providing a measure of uncertainty. This uncertainty flags ambiguous cases for human review, improving reliability when the system encounters novel data outside its training distribution. Additionally, integrating UQ within an ensemble framework enhances interpretability, as uncertainty estimates can be analyzed alongside visual explanations generated by XGrad-CAM, ensuring that both the confidence and rationale behind predictions are transparent to clinicians.

\subsection{Research Gaps}

Existing segmentation methods rely heavily on pixel-level annotations or bounding boxes, increasing the time, cost, and expertise required for data preparation. Explainability techniques like CAM and Grad-CAM generate heatmaps but cannot produce pixel-level binary masks. At the same time, MDVA-GAN, despite addressing this limitation, suffers from instability and misclassification of unseen data. Open-set detection remains challenging, as models often fail to identify novel inputs outside the training distribution. Although UIOS estimates uncertainty for open-set scenarios, its dependence on specific annotations and manual thresholding limits its generalizability. Moreover, few methods integrate uncertainty quantification and visual explainability within the same framework, reducing reliability and transparency. This study addresses these gaps by developing a weakly supervised approach that uses only image-level labels to generate pixel-level annotations. By integrating Monte Carlo Dropout within an ensemble of ResNet50, EfficientNet, and DenseNet models, the system estimates prediction uncertainty to improve reliability in open-set scenarios. Additionally, XGrad-CAM enhances interpretability by providing visual explanations, ensuring both confidence and reasoning behind predictions are transparent and accessible.

\section{Auto Annotation eXplainable (AXX) Model}

The Auto Annotation eXplainable (AAX) Model integrates three pre-trained deep learning models (ResNet50, DenseNet, and EfficientNet) enhanced with Monte Carlo Dropout for uncertainty estimation and XGrad-CAM for visual explainability. This ensemble framework is designed to classify chest X-ray images as "Diseased" or "Healthy" while providing robust prediction confidence and pixel-level annotations using only image-level labels. Combining these models ensures complementary feature extraction: ResNet50 captures low-level patterns, EfficientNet balances efficiency and accuracy, and DenseNet leverages dense connections to improve feature propagation.

The methodology is structured into three phases: the Training Phase, Inference and Output, and Decision Protocol, each aligning with one of the three main contributions: (1) ensemble learning for robust predictions, (2) weakly supervised pixel-level annotations using XGrad-CAM, and (3) uncertainty quantification for open-set detection.

\begin{figure}[!h]
    \centering
    \includegraphics[width=\columnwidth, trim={0.5cm 0.5cm 0.5cm 0.5cm}, clip]{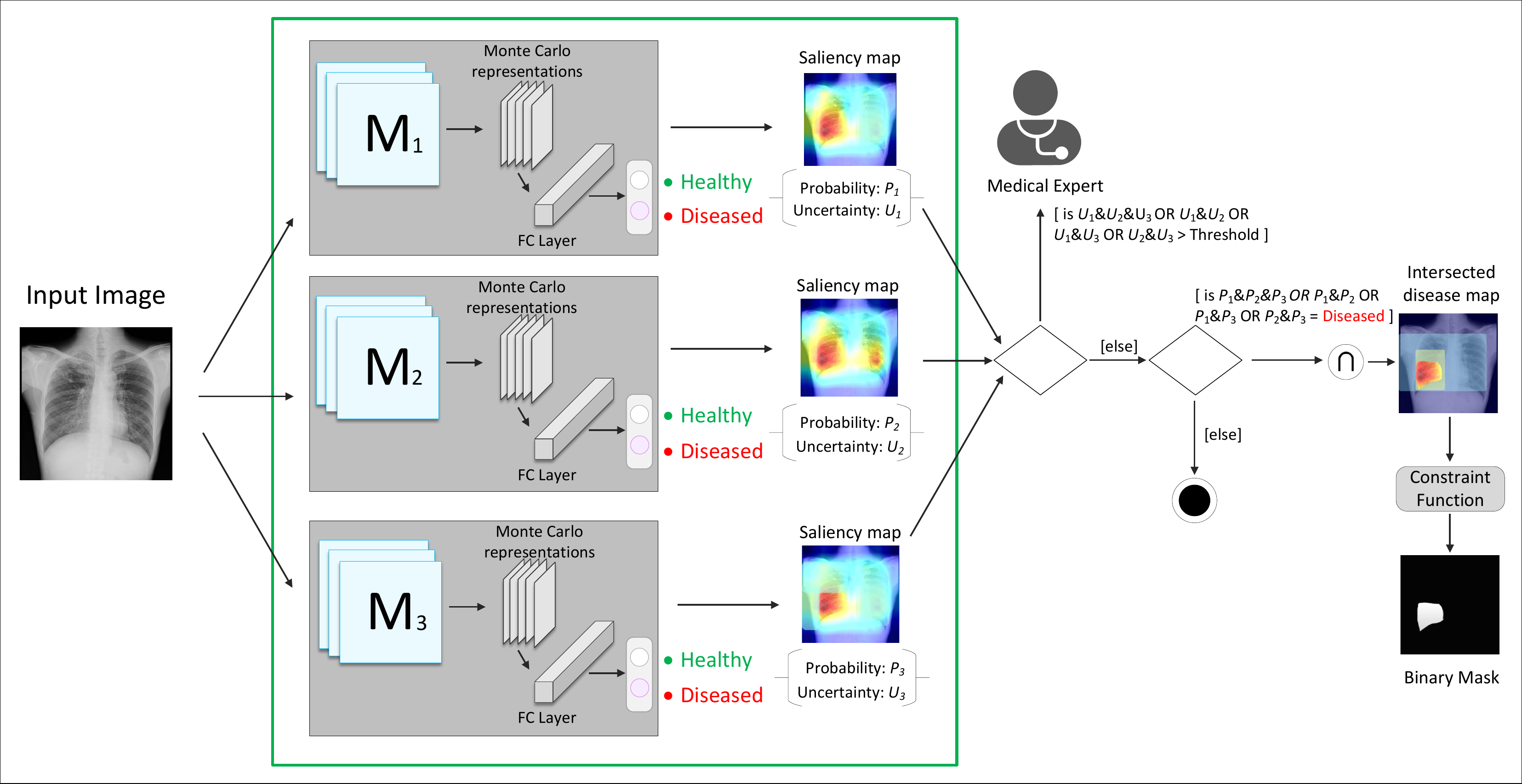}
    \caption{\textbf{Architecture of the Auto Annotation eXplainable (AAX) Model.} 
    The pipeline integrates ResNet50, EfficientNet, and DenseNet with Monte Carlo 
    Dropout and XGrad-CAM. Images with high uncertainty in at least two models 
    are flagged for expert review, while confident predictions are classified as 
    "Healthy" or "Diseased." For diseased cases, intersected saliency maps 
    generate a binary mask highlighting disease-specific regions.}
    \label{fig:proposed_methodology}
\end{figure}

Figure \ref{fig:proposed_methodology} provides a visual representation of the AAX architecture, illustrating the flow from input image analysis to final classification and annotation. This diagram highlights the role of each model, the generation of saliency maps, and the integration of uncertainty scores into the decision-making process.

\subsection{Training Phase}

Each model \(M_i\) in the ensemble is independently trained on a dataset \(D = \{(x_j, y_j)\}\), where \(x_j\) is an input image and \(y_j\) is the corresponding image-level label. The objective is to minimize a composite loss function \(L\) that balances classification accuracy and prediction uncertainty, with Monte Carlo Dropout encouraging diverse predictions across stochastic forward passes. Formally, the loss function is:

\[
L(M_i, D) = L_{\text{cls}}(M_i, D) + \lambda L_{\text{dropout}}(M_i, D)
\]

where \(L_{\text{cls}}\) is the cross-entropy classification loss, \(L_{\text{dropout}}\) is the variance of predictions across stochastic passes, and \(\lambda\) controls the balance between these components. This formulation enables each model to not only classify input images but also quantify the uncertainty associated with its predictions.

\subsection{Inference and Output}
During inference, an input image \(x\) is processed concurrently through all three models. Each model \(M_i\) performs multiple stochastic forward passes using Monte Carlo Dropout to produce three outputs:

\begin{itemize}
    \item \(p_i(x)\): The probability of the image being "Diseased" (derived from the softmax layer).
    \item \(u_i(x)\): The uncertainty score is calculated as the standard deviation of the predictive distribution.
    \item \(S_i(x)\): The saliency map generated using XGrad-CAM, highlighting regions that most influence the model’s prediction.
\end{itemize}

These outputs are then aggregated to form the final prediction and pixel-level annotation, ensuring that the ensemble’s collective judgment is both robust and interpretable.

\subsection{Decision Protocol}

The decision-making process integrates both consensus among models and uncertainty thresholding to improve reliability, particularly in open-set scenarios. The protocol is as follows:

\noindent
\textbf{(a) Uncertainty Thresholding:}
    \begin{itemize}
        \item An uncertainty threshold \(\theta\) is defined to identify ambiguous or novel cases. If at least two models produce uncertainty scores \(u_i(x)\) above \(\theta\), the image is flagged for expert review:
        \[
        \sum_{i=1}^{3} \mathbb{I}[u_i(x) > \theta] \geq 2
        \]
    \end{itemize}

\noindent
\textbf{(b) Consensus-Based Prediction:}
    \begin{itemize}
        \item If uncertainty scores are below \(\theta\), the ensemble classifies the image based on consensus. For a "Diseased" classification, at least two models must output probabilities \(p_i(x) > 0.5\) with low uncertainty:
        \[
        \sum_{i=1}^{3} \mathbb{I}[p_{\text{Diseased}, i}(x) > 0.5 \wedge u_i(x) < \theta] \geq 2
        \]
        
        \item The saliency maps of the agreeing models are then intersected to generate a binary mask highlighting disease-specific features:
        \[
        S_{\text{int}}(x) = \bigcap_{\{i \mid p_{\text{Diseased}, i}(x) > 0.5 \wedge u_i(x) < \theta\}} S_i(x)
        \]
        
        \item If at least two models classify the image as "Healthy" with low uncertainty, the image is labeled as "Healthy" without generating a saliency map:
        \[
        \sum_{i=1}^{3} \mathbb{I}[p_{\text{Healthy}, i}(x) > 0.5 \wedge u_i(x) < \theta] \geq 2
        \]
    \end{itemize}

This protocol ensures that the system only provides pixel-level annotations when predictions are confident and consistent across multiple models, reducing the risk of erroneous annotations while maintaining interpretability.

\subsection{Pseudocode}

The operational framework of the AAX model is summarized in Algorithm 1, which outlines the process from input image analysis to final classification and annotation. The pseudocode ensures reproducibility and provides a clear reference for implementing the model.

\begin{algorithm}[H] \caption{AXX Model} \label{alg:aax} \begin{algorithmic}[1] \REQUIRE $X$: Set of input images, $\theta$: Uncertainty threshold, $M = {M_1, M_2, M_3}$: Set of pre-trained models
\ENSURE Class predictions, uncertainty values, saliency maps, binary masks

\FOR{each image $x \in X$}
\FOR{each model $M_i \in M$}
\STATE Perform $K$ stochastic forward passes using MC Dropout
\STATE Calculate mean probability $p_i(x)$ and uncertainty $u_i(x)$
\STATE Generate saliency map $S_i(x)$ using XGrad-CAM
\ENDFOR
\IF{$\sum_{i=1}^{3} \mathbf{1}_{u_i(x) > \theta} \geq 2$}  
    \STATE Flag $x$ for expert review  
\ELSE  
    \IF{$\sum_{i=1}^{3} \mathbf{1}_{p_{\text{Diseased},i}(x) > 0.5 \land u_i(x) < \theta} \geq 2$}  
        \STATE Classify $x$ as "Diseased"  
        \STATE $S_{\text{int}}(x) \leftarrow \bigcap_{\{i | p_i(x) > 0.5 \land u_i(x) < \theta\}} S_i(x)$  
        \STATE Convert $S_{\text{int}}(x)$ to a binary mask  
    \ELSE  
        \STATE Classify $x$ as "Healthy"  
    \ENDIF  
\ENDIF  
\ENDFOR
\end{algorithmic}
\end{algorithm}

\subsection{Summary}

The AAX Model integrates ResNet50, DenseNet, and EfficientNet within an ensemble framework to reduce model variability and enhance classification accuracy. Each model, equipped with Monte Carlo Dropout and XGrad-CAM, processes input images to generate class probabilities, uncertainty scores, and saliency maps. This design enables weakly supervised pixel-level annotations using only image-level labels, eliminating the need for manual pixel annotations. By intersecting saliency maps from models that classify an image as "Diseased" with low uncertainty, the system produces a binary mask highlighting disease-specific regions. In cases where at least two models exhibit high uncertainty, the image is flagged for expert review, ensuring reliable performance in open-set scenarios. This approach directly addresses the identified research gaps, combining ensemble learning for robust predictions, weakly supervised annotation with XGrad-CAM, and uncertainty quantification through Monte Carlo Dropout to enhance reliability and interoperability.

\section{Experiments}

To evaluate the performance, reliability, and generalizability of the AXX Model, we conducted experiments on both medical and general image segmentation datasets. These experiments assess the model’s ability to generate accurate pixel-level annotations using only image-level labels, as well as its capacity to quantify uncertainty and flag ambiguous cases. Using two distinct datasets we aim to highlight the model’s versatility and robustness across different application domains.

\subsection{Experimental Setup}

The AXX Model was evaluated using two datasets: 
\begin{itemize}
    \item \textbf{TBX11K dataset for tuberculosis detection \cite{ref42}.} It is a key resource for tuberculosis (TB) detection, comprising 11,200 high-resolution chest X-ray images (512 × 512 pixels) across four categories: healthy, active TB, latent TB, and non-TB unhealthy. Its diverse cases support robust diagnostic model development. Training primarily focuses on distinguishing 'healthy' from 'TB' using weakly supervised learning, eliminating the need for bounding boxes and allowing models to learn from global image features. 
    \item \textbf{Fire Segmentation dataset for general image segmentation \cite{ref55}.} It comprises video frames from YouTube, categorized into $Fire$ (27,460 images with visible fire pixels of varying intensities and spreads) and $Not_{Fire}$ (11,392 images without fire). All images are standardized to 256 × 256 pixels, ensuring consistency for training and evaluating fire detection models.
\end{itemize}

Each model (ResNet50, DenseNet, and EfficientNet) was trained using the Adam optimizer with a learning rate of 0.0001, a batch size of 16, and a maximum of 50 epochs. Early stopping was applied, terminating training if validation loss did not improve for five consecutive epochs. Monte Carlo Dropout was implemented during both training and inference, with XX stochastic forward passes performed per image to estimate prediction uncertainty. The uncertainty threshold  $\theta=0.1$ was determined empirically using validation data to balance sensitivity and specificity in detecting ambiguous cases.

\subsection{Evaluation Metrics}

We used the following metrics to evaluate model performance:

\begin{itemize}
    \item \textbf{Accuracy:} The ratio of correctly classified samples to the total number of samples, measuring overall classification performance.
    \[
    \text{Accuracy} = \frac{TP + TN}{TP + TN + FP + FN}
    \]

    \item \textbf{Precision:} The ratio of correctly predicted positive pixels to the total pixels predicted as positive indicates prediction reliability.
    \[
    \text{Precision} = \frac{TP}{TP + FP}
    \]

    \item \textbf{Recall:} The ratio of correctly predicted positive pixels to the total actual positive pixels, assessing the model’s ability to capture all positives.
    \[
    \text{Recall} = \frac{TP}{TP + FN}
    \]

    \item \textbf{F1-Score:} The harmonic mean of precision and recall, balancing both metrics in binary classification tasks.
    \[
    \text{F1-Score} = \frac{2 \times \text{Precision} \times \text{Recall}}{\text{Precision} + \text{Recall}}
    \]

    \item \textbf{Intersection over Union (IoU):} The ratio of correctly predicted pixels to the union of predicted and actual pixels, evaluating pixel-level segmentation accuracy.
    \[
\text{IoU} = \frac{\text{Area of Overlap}}{\text{Area of Union}}
\]

\end{itemize}

\section{Results}

\subsection{Quantitative Results}

The quantitative performance of the AAX Model is evaluated on two datasets: Fire and TBX11K. Results are presented separately for each dataset, with tables summarizing model accuracy and pixel-level annotation performance using IoU.\\

\noindent
\textbf{Fire Dataset}. 
Table \ref{tab:acc_performance_fire} shows the classification accuracy of the AAX model compared to baseline methods, while Table \ref{tab:iou_performance_fire} reports IoU scores, reflecting pixel-level annotation performance. The AAX model achieved the highest accuracy and IoU, demonstrating the benefits of ensemble learning and weakly supervised pixel-level annotations. The low uncertainty flag rate indicates that the model maintained high confidence across most predictions while effectively identifying ambiguous cases.

\input{tab_res1} 

\input{tab_res2} 

\noindent
\textbf{TBX11K Dataset}. Table \ref{tab:model_accuracy_tbx} presents the classification accuracy for the TBX11K dataset, and Table \ref{tab:iou_performance_tbx} reports IoU scores, evaluating the model’s ability to localize tuberculosis-related abnormalities. Despite the complexity of medical image annotation, the AAX model outperformed baseline methods, highlighting the effectiveness of using XGrad-CAM for weakly supervised pixel-level annotations. The uncertainty flag rate was higher than the Fire dataset, indicating that the model correctly identified more ambiguous or novel cases, which is essential in healthcare applications.

\input{tab_res3} 

\input{tab_res4} 

\noindent
\textbf{Summary.} Across both datasets, the AAX model consistently achieved higher accuracy and IoU than individual models and MDVA-GAN. Combining ensemble learning, weakly supervised annotations, and uncertainty quantification contributed to this performance, validating the model’s generalizability and reliability. Further interpretation of these results and their implications is provided in the Discussion section.

\subsection{Qualitative Results}
The qualitative performance of the AXX Model is illustrated using visual comparisons from both the TBX11K and Fire datasets. Each example shows the input image, ground truth mask, predictions from baseline methods, and the AAX model’s pixel-level annotations. Saliency maps generated using XGrad-CAM highlight regions most influential to the model’s predictions, demonstrating the interpretability of the annotations.\\

\noindent
\textbf{(a) Fire Dataset.} 

The Fire Segmentation dataset was selected for its visually interpretable data, making it accessible to both experts and non-experts. This dataset tests the AAX model’s ability to focus on class-specific features, such as fire pixels, demonstrating its versatility beyond medical imaging.

\begin{figure}[!h]
\resizebox{\columnwidth}{!}{
\includegraphics{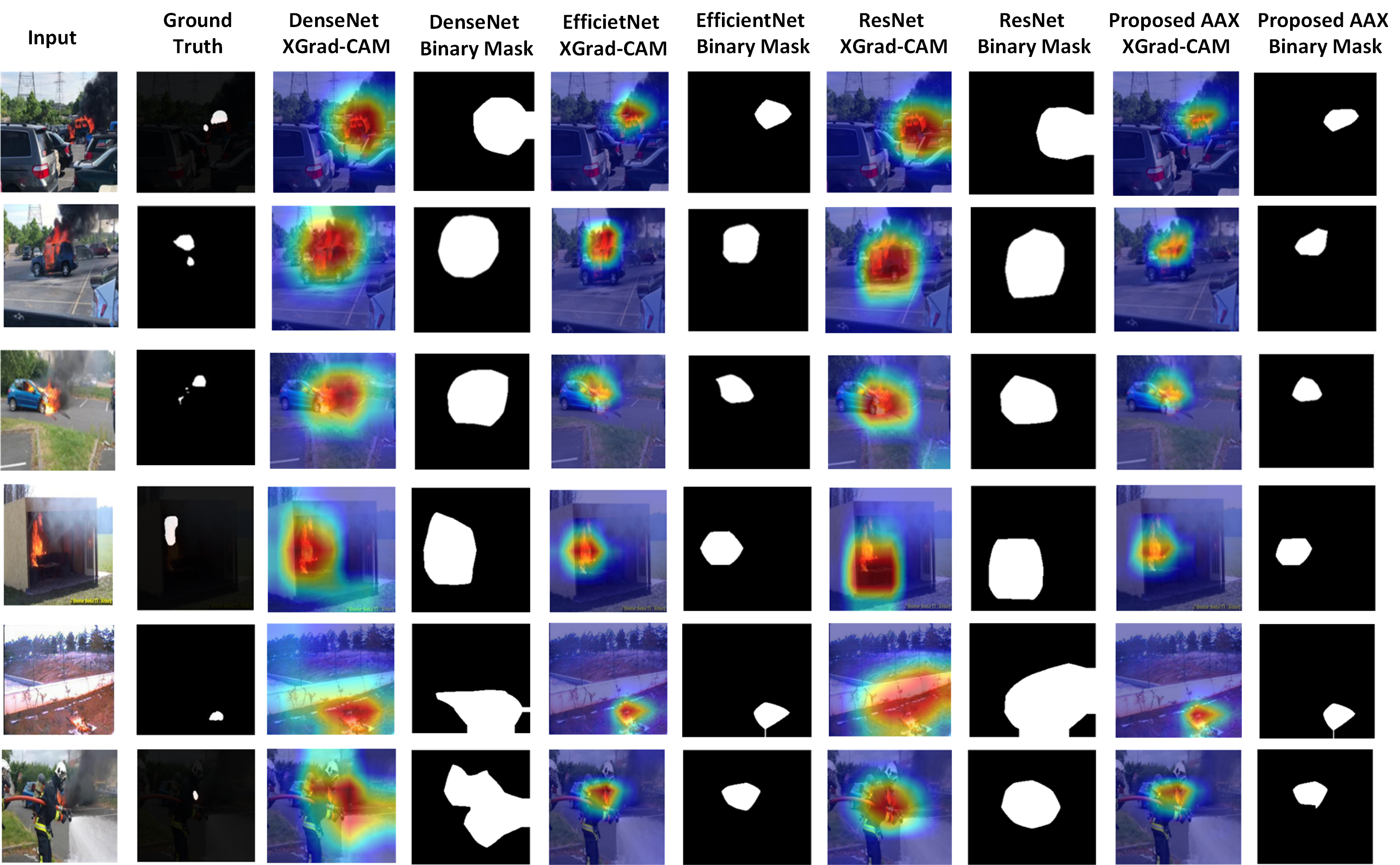} 
}
\caption{\textbf{Comparative Results of AAX Model and Baseline Methods on Fire Dataset.} Visual comparison of input images, ground truth masks, and model outputs from DenseNet, EfficientNet, ResNet50, and the proposed AAX model. Columns show XGrad-CAM saliency maps and binary masks for each model. The AAX model’s binary masks align more closely with ground truth annotations, demonstrating improved feature localization and pixel-level annotation accuracy.}
\label{comparitive_on_fire}
\end{figure}

Figure \ref{comparitive_on_fire} compares the AAX model with ResNet50, EfficientNet, and DenseNet, respectively. Each figure shows the input image, ground truth mask, saliency maps generated using XGrad-CAM, and binary masks derived from each model’s predictions. The AAX model’s binary masks consistently align more closely with the ground truth, demonstrating superior pixel-level annotation accuracy. 




\noindent
\textbf{(b) TBX11K Dataset.}

The TBX11K dataset, designed for tuberculosis detection in chest X-rays, includes bounding-box annotations but lacks pixel-level labels. Despite being trained using only image-level labels, the AAX model successfully generates pixel-level masks, enhancing the dataset’s utility for segmentation tasks and disease localization.

\begin{figure}[!h]
\resizebox{\columnwidth}{!}{
\includegraphics{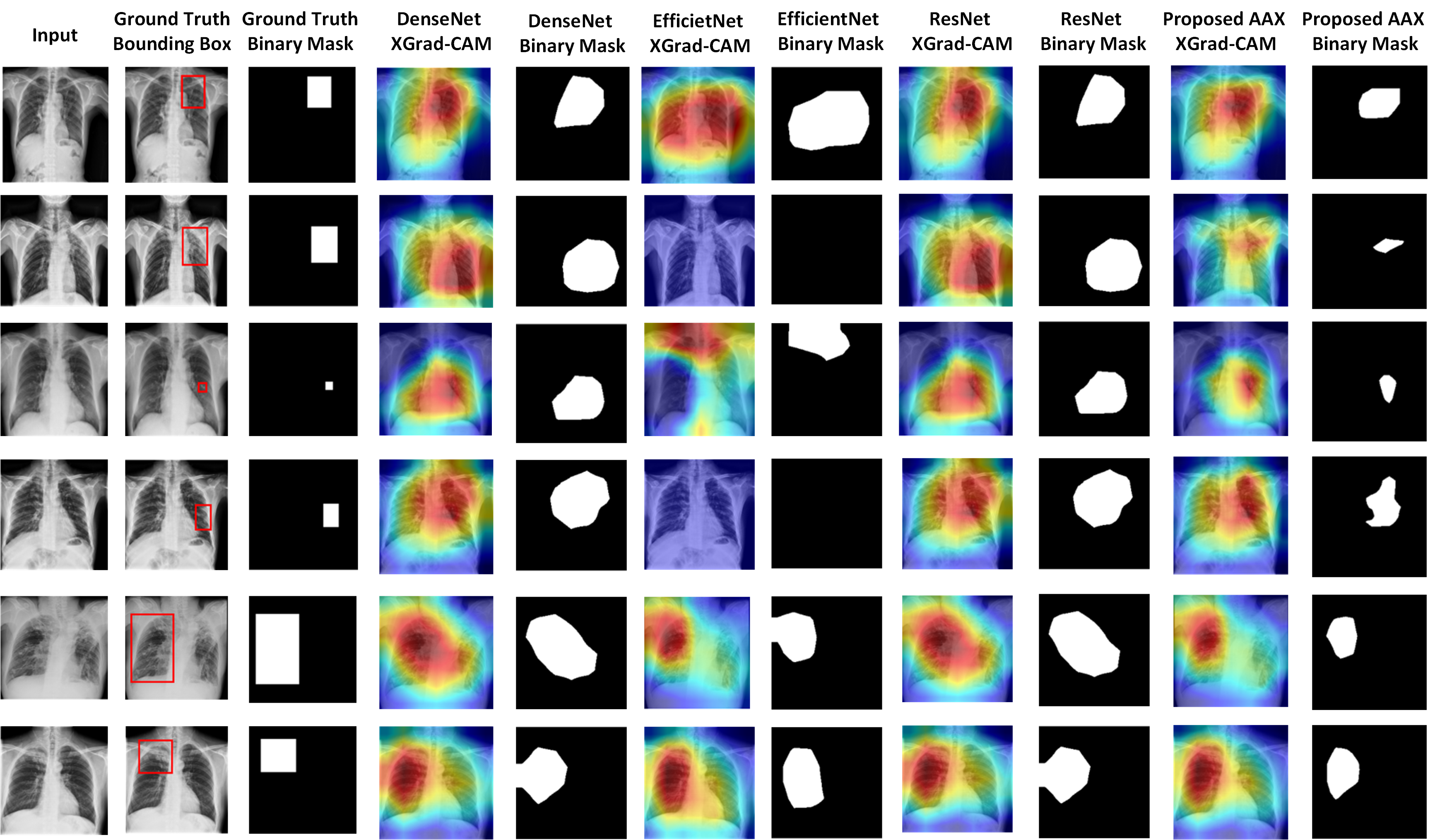} 
}
\caption{\textbf{Comparative Results of AAX Model and Baseline Methods on TBX11K Dataset.} Visual comparison of chest X-ray inputs, ground truth annotations (bounding boxes and binary masks), and model outputs from DenseNet, EfficientNet, ResNet50, and the proposed AAX model. The AAX model's binary masks align more closely with ground truth annotations, demonstrating improved pixel-level localization despite being trained using only image-level labels.}
\label{tbx_general}
\end{figure}

Figure \ref{tbx_general} compares the proposed AXX model with ResNet50, EfficientNet, and DenseNet, respectively. Each figure includes input images, ground truth bounding boxes, binary masks derived from these boxes, and saliency maps from each model. The AAX model's binary masks demonstrate greater alignment with the ground truth, accurately localizing tuberculosis regions despite the lack of pixel-level training data. 




\noindent
\textbf{Summary.} The qualitative results demonstrate that the AAX model produces more accurate pixel-level annotations compared to baseline methods, aligning closely with ground truth masks despite being trained using only image-level labels. The ensemble approach enhances feature localization, while uncertainty quantification ensures that ambiguous cases are flagged for expert review. These visual comparisons validate the model’s three main contributions, showcasing its effectiveness in medical and general image segmentation tasks.

\section{Discussion}

\noindent (1) \textbf{Ensemble Learning Improves Prediction Accuracy:} The AAX model outperformed individual ResNet50, DenseNet, and EfficientNet models, confirming that ensemble learning reduces model variability and enhances classification accuracy. By combining the diverse feature extraction capabilities of the three models, the ensemble achieved higher accuracy and IoU scores, demonstrating the benefit of aggregating predictions from complementary architectures. This improvement was consistent across both medical (TBX11K) and general (Fire) image segmentation tasks, highlighting the robustness of the ensemble approach.\\

\noindent (2) \textbf{Weakly Supervised Annotations Align with Ground Truth:} Despite being trained using only image-level labels, the AAX model produced pixel-level masks that closely matched ground truth annotations. By leveraging XGrad-CAM to generate saliency maps and intersecting these maps across models with consistent predictions, the system effectively localized disease-specific regions. This weakly supervised approach eliminates the need for manual pixel-level annotations, reducing data preparation costs and making the model more scalable for real-world applications.\\

\noindent (3) \textbf{Uncertainty Quantification Enhances Open-Set Detection:} Monte Carlo Dropout enabled the model to estimate prediction uncertainty, improving reliability in open-set scenarios. By flagging highly uncertain images for expert review, the system reduced the risk of misclassification for novel or ambiguous cases. The relatively low uncertainty flag rates of 8.2\% for TBX11K and 6.5\% for Fire indicate that the model maintained high confidence in most predictions while still identifying challenging cases. This balance ensures both diagnostic efficiency and safety, aligning with clinical requirements for reliable AI-assisted decision-making.\\

\noindent (4) \textbf{Generalizability Across Domains Demonstrates Model Robustness:} The AAX model’s high performance on TBX11K and Fire datasets demonstrates its generalizability beyond medical imaging. The consistent improvement in accuracy and IoU across different domains validates the effectiveness of the ensemble approach and highlights the potential for broader applications, including industrial inspection, environmental monitoring, and object detection. This versatility positions the AAX model as a reliable solution for diverse segmentation tasks where pixel-level annotations are unavailable.\\

\noindent (5) \textbf{Visual Interpretability Builds Trust in Automated Predictions:} The saliency maps generated using XGrad-CAM provided visual explanations aligned with clinically relevant regions, enhancing the interpretability of model predictions. By intersecting saliency maps from models with consistent classifications, the system highlighted critical features with greater precision, reducing false positives and improving diagnostic clarity. The visual alignment between predicted masks and ground truth annotations reinforces the model’s reliability, supporting its integration into healthcare workflows where explainability is essential for clinical trust.\\

\noindent (6) \textbf{Open-Set Detection Reduces the Risk of Misdiagnosis:} The model’s ability to flag uncertain cases for expert review is crucial for ensuring reliability in real-world applications. By setting an uncertainty threshold of \(\theta = 0.1\), the system effectively identified edge cases with subtle abnormalities or overlapping features. This mechanism prevents the model from making overconfident predictions on novel or ambiguous inputs, reducing the risk of misdiagnosis and supporting a collaborative workflow where AI assists rather than replaces human expertise.\\

\noindent (7) \textbf{Experimental Results Validate Key Contributions:} The experimental results directly support the three main contributions of this study. Ensemble learning improved prediction accuracy and consistency, weakly supervised pixel-level annotations generated interpretable masks without manual labels, and uncertainty quantification enabled reliable open-set detection. Together, these contributions address critical challenges in medical image analysis and demonstrate the potential of explainable AI to enhance diagnostic accuracy and efficiency.

\section{Limitation and Challenges}

While the proposed framework demonstrates strong performance in single-label classification tasks, several challenges remain, particularly in more complex scenarios. The system’s performance in multi-label classification is limited, especially when known and unknown classes coexist. Although the uncertainty module effectively identifies novel patterns, the simultaneous presence of multiple labels can cause classification ambiguity, reducing prediction accuracy.

Additionally, the ensemble-based decision-making relies on the collective performance of ResNet50, EfficientNet, and DenseNet. If two out of the three models fail to converge adequately during training, the system’s consensus mechanism is compromised, leading to decreased stability and predictive accuracy. This dependency highlights the need for robust training processes to ensure consistent performance across all ensemble members.

Addressing these limitations in future work will involve enhancing the model's multi-label classification capabilities, refining the uncertainty module to better differentiate between known and unknown classes, and optimizing the ensemble framework to reduce reliance on individual model performance.

\section{Ethical Considerations}

The development and deployment of the AAX Model are guided by principles of transparency, accountability, and fairness, particularly in high-stakes domains like medical imaging. By integrating XGrad-CAM for visual explanations, the model enhances interpretability, allowing users to understand and validate its predictions, thus fostering trust in AI-assisted decision-making. Monte Carlo Dropout for uncertainty quantification ensures that ambiguous cases are flagged for expert review, promoting human oversight and reducing the risk of misdiagnosis.  

To mitigate bias, future work will focus on training and validating the model using diverse datasets to ensure consistent performance across different populations. Additionally, the system's reliance on image-level labels reduces the need for detailed annotations, minimizing the risk of exposing sensitive information during data labeling. The AAX model is intended \textit{to support, not replace, human expertise, ensuring that AI serves as an assistive tool that enhances diagnostic accuracy and efficiency while maintaining ethical standards in healthcare and beyond}.

\section{Conclusions and Future Work}

This study introduced an automated data annotation system that improves classification accuracy, pixel-level annotations, and prediction reliability. \textit{(1) Ensemble learning enhances accuracy and consistency:} Combining ResNet50, EfficientNet, and DenseNet reduced variability, achieving higher accuracy and IoU scores on both TBX11K and Fire datasets; \textit{(2) Weakly supervised learning enables pixel-level annotations using only image-level labels:} By intersecting XGrad-CAM saliency maps, the system generated masks closely aligned with ground truth annotations, reducing annotation costs. \textit{(3) Uncertainty quantification improves reliability in open-set scenarios:} Monte Carlo Dropout flagged ambiguous cases for expert review, maintaining high confidence while detecting novel inputs. Future work will focus on incremental learning to adapt to new data and advanced open-set detection methods to identify unseen classes, enhancing scalability and robustness across diverse applications.

\section{Source Code}

The source code for all experiments and the implementation of the Auto Annotation eXplainable Model is available at \url{https://github.com/basmakhan11/Auto-Annotation-eXplainable-AAX-Model)}.

\section{Acknowledgments}
The work reported in this article was partially supported under the auspices of the UNESCO Chair on AI \& VR by national funds through Funda\c{c}\~{a}o para a Ci\^{e}ncia e a Tecnologia with references DOI:10.54499/\-UIDB/\-50021/2020,
DOI:10.54499/DL57/2016/CP1368/\-CT0002
and 2022.09212.PTDC (XAVIER project).\\

\end{document}

%% file: tab_res1.tex
\begin{table}[!h]
\centering
\caption{\textbf{Fire Dataset. Comparison of the AAX Model with State-of-the-Art Methods.} The table presents the average accuracy of the proposed AAX model compared to baseline and state-of-the-art methods. The AAX model achieves the highest accuracy, demonstrating the effectiveness of ensemble learning and weakly supervised annotations.}
\label{tab:acc_performance_fire}
\begin{tabular}{@{}lc@{}}
\toprule
\textbf{Methodology} & \textbf{Average Accuracy (\%)} \\ \midrule
SGD \cite{ref43}            & 92.27                         \\
AlexNet \cite{ref44}        & 94.8                          \\
DBN \cite{ref45}           & 95.0                          \\
Inceptionv3 \cite{ref46}     & 93.6                          \\
ForestResNet \cite{ref47}    & 92.0                          \\
VGG19 \cite{ref48}        & 94.0                          \\
ResNet50             & 93.7                          \\
EfficientNet         & 95.1                          \\
DenseNet             & 92.6                          \\
\textbf{Proposed AAX}   & \textbf{96.4}                          \\ \bottomrule
\end{tabular}
\end{table}

%% file: tab_res2.tex
\begin{table}[!h]
\centering
\caption{\textbf{Fire Dataset. IoU Scores.} 
The table compares the IoU scores of individual models using XGrad-CAM and Monte Carlo Dropout with the proposed AAX model. The AAX model achieves the highest IoU, demonstrating improved pixel-level annotation accuracy through ensemble learning.}
\label{tab:iou_performance_fire}
\begin{tabular}{@{}lc@{}}
\toprule
\textbf{Methodology} & \textbf{IoU (\%)} \\ \midrule
ResNet50 with XGrad-CAM and Monte Carlo & 51.2 \\
DenseNet with XGrad-CAM and Monte Carlo & 49.6 \\
EfficientNet with XGrad-CAM and Monte Carlo & 55.1 \\
Proposed AAX Model & 64.7 \\ \bottomrule
\end{tabular}
\end{table}

%% file: tab_res3.tex
\begin{table}[!h]
\centering
\caption{\textbf{TBX11K. Comparison of the AAX Model with State-of-the-Art Methods.}
The table presents the average accuracy of the proposed AAX model compared to baseline and state-of-the-art methods. The AAX model achieves the highest accuracy, demonstrating the effectiveness of ensemble learning and uncertainty quantification in medical image classification.}
\label{tab:model_accuracy_tbx}
\begin{tabular}{@{}lc@{}}
\toprule
\textbf{Methodology} & \textbf{Average Accuracy (\%)} \\ \midrule
ResNet50 without Monte Carlo Dropout \cite{ref49} & 92.61 \\
CNN and Grad-CAM \cite{ref50} & 92.5 \\
UIOS with Dirichlet distribution \cite{ref38} & 88.27 \\
SSD \cite{ref51} & 84.7 \\
RetinaNet \cite{ref52} & 87.45 \\
Faster R-CNN \cite{ref53} & 89.7 \\
FCOS \cite{ref54} & 88.9 \\
ResNet50 with Monte Carlo Dropout & 89.1 \\
EfficientNet with Monte Carlo Dropout & 90.3 \\
DenseNet with Monte Carlo Dropout & 88.7 \\
\textbf{Proposed AAX} & \textbf{93.04} \\ \bottomrule
\end{tabular}
\end{table}

%% file: tab_res4.tex
\begin{table}[!h]
\centering
\caption{\textbf{TBX11K. IoU Scores.} The table compares the IoU scores of the proposed AAX model with baseline and state-of-the-art methods. The AAX model achieves the highest IoU, demonstrating improved pixel-level annotation accuracy using ensemble learning and weakly supervised segmentation.}
\label{tab:iou_performance_tbx}
\begin{tabular}{@{}lc@{}}
\toprule
\textbf{Methodology} & \textbf{IoU (\%)} \\ \midrule
MDVA-GAN \cite{ref30} & 25.40 \\
VA-GAN \cite{ref24} & 29.15 \\
Pre-trained ResNet50 (with XGrad-CAM) & 16.1 \\
DenseNet (with XGrad-CAM) & 15.0 \\
EfficientNet (with XGrad-CAM) & 23.3 \\
\textbf{Proposed AAX} & \textbf{36.07} \\ \bottomrule
\end{tabular}
\end{table}